\documentclass{article}
\usepackage{Odyssey2016,cite,amssymb,amsmath,epsfig,url,graphicx,multirow,threeparttable,float,balance,array,bm}
\ninept
\title{Spoofing detection under noisy conditions: \\a preliminary investigation and an initial database}

\makeatletter
\def\name#1{\gdef\@name{#1\\}}
      
\name{\em Xiaohai Tian$^{1,2}$, Zhizheng Wu$^3$, Xiong Xiao$^4$,  Eng Siong Chng$^{1,2,4}$ and Haizhou Li$^{1,5}$}
\address{$^1$School of Computer Engineering, Nanyang Technological University (NTU),Singapore \\
$^2$Joint NTU-UBC Research Center of Excellence in Active Living for the Elderly, NTU, Singapore \\
$^3$The Center for Speech Technology Research, University of Edinburgh, United Kingdom \\
$^4$Temasek Laboratories, NTU, Singapore \\
$^5$Human Language Technology Department, Institute for Infocomm Research, Singapore \\
{\small \tt \{xhtian, xiaoxiong, aseschng\}@ntu.edu.sg, \tt zhizheng.wu@ed.ac.uk, \tt hli@i2r.a-star.edu.sg}
}

\begin{document}
\maketitle
\begin{abstract}
Spoofing detection for automatic speaker verification (ASV), which is to discriminate between live speech and attacks, has received increasing attentions recently.
However, all the previous studies have been done on the clean data without significant additive noise. 
To simulate the real-life scenarios, we perform a preliminary investigation of spoofing detection under additive noisy conditions, and also describe an initial database for this task.
The noisy database is based on the ASVspoof challenge 2015 database and generated by artificially adding background noises at different signal-to-noise ratios (SNRs). Five different additive noises are included.
Our preliminary results show that using the model trained from clean data, the system performance degrades significantly in noisy conditions. Phase-based feature is more noise robust than magnitude-based features. And the systems perform significantly differ under different noise scenarios.

\end{abstract}

\section{Introduction}

Recently, automatic speaker verification (ASV) has been significantly advanced to the point of mass-market adoption~\cite{lee2013sltc,khitrov2013talking,beranek2013voice,meng2015surveying}. However, most of current ASV systems assume human voices, and there are concerns that whether the systems can still achieve robust performance in the face of diverse spoofing attacks. A spoofing attack is that an attacker attempts to manipulate an ASV result for a target genuine speaker to obtain access permission. A significant amount of evidences have confirmed the vulnerability of current state-of-the-art ASV systems under spoofing attacks as reviewed in~\cite{wu2014survey}. This has led to the active development of spoofing countermeasures, also called spoofing detection, that is to discriminate human and spoofed speech. 

In the past several years, spoofing detection for speaker recognition has been studied on a variety of diverse datasets. In~\cite{6205335,wu2013synthetic}, the Wall Street Journal (WSJ) corpus was used to assess countermeasures for speech synthesis attacks. In~\cite{wu2014replay}, the publicly available RSR2015 corpus was used to evaluate spoofing detection for replay attacks. In~\cite{de2012synthetic, sanchez2015toward}, synthetic speech from the Blizzard challenge~\cite{king2014measuring} was used for speech synthesis spoofing detection. In~\cite{wu2016antispoofing}, a recently released spoofing and anti-spoofing (SAS) corpus as a standard spoofing database was used to assess speech synthesis and voice conversion spoofing countermeasures. We note that WSJ, SAS and Blizzard challenge databases were recorded by high-quality microphones in sound-proofing environment, while the RSR2015 corpus was recorded by multiple mobile devices in a quiet office room. All these databases do not have any significant channel and/or additive noise. These databases allow us to focus on spoofing effects but do not simulate practical scenarios of ASV applications. 

There are also some studies that use data with channel noise. The National Institute of Standards and Technology (NIST) Speaker Recognition Evaluation (SRE) 2006 database which has significant telephone channel noise was used to assess voice conversion spoofing countermeasures in~\cite{wu2012detecting,wu2012study,alegre2013conversion,khoury2014introducing,Sizov2015joint}. In~\cite{ergunay2015vulnerability}, a so-called AVspoof database includes replay, speech synthesis and voice conversion spoofing attacks to simulate realistic scenarios, which re-recorded synthetic or voice-converted speech using multiple mobile devices.

In general, all the databases used in the past spoofing detection studies do not consider additive noise, even the recent standard spoofing detection databases: SAS\footnote{SAS corpus is available at:~\url{http://dx.doi.org/10.7488/ds/252}} and ASVspoof 2015 challenge\footnote{ASVspoof 2015 corpus is available at:~\url{http://dx.doi.org/10.7488/ds/298}} databases. However, in practical scenarios, it is hard to avoid additive and/or channel noise. Hence, another concern for ASV deployment arises that whether currently developed spoofing detection algorithms/systems are still effective under noisy conditions. 

In this work, we focus on spoofing detection under additive noisy conditions. We perform a preliminary investigation of spoofing detection under noisy conditions using current state-of-the-art countermeasure techniques; and also briefly introduce an initial database we built for the task\footnote{The noisy database will be publicly-available for free under a CC-BY license.}. In general, we aim to answer the following questions:
\begin{itemize}
    \item Do current state-of-the-art spoofing detection algorithms work well under additive noisy conditions?
    \item How additive noises affect the spoofing detection performance?
    \item What kind of noise is more serious to degrade the performance of spoofing detection algorithms?
\end{itemize}
We believe better understanding of above questions and the noisy database will drive the development of generalised and noise robust spoofing detection algorithms.

\section{Noisy Database}
\label{sec:noise_database}

In order to represent the practical application scenarios for spoofing detection, we attempt to design a database in additive noisy environments based on the ASVspoof 2015 challenge database~\cite{wu15asvspoof}.

ASVspoof database is a spoofing and anti-spoofing database, consisting of both genuine (human speech) and ten types of spoofed speech (named as S1-S10 in ASVspoof 2015 challenge) implemented by three speech synthesis and seven voice conversion spoofing algorithms. The ASVspoof database contains three subsets, including training, development and evaluation sets. The training and development sets only contain \emph{known} attacks (S1-S5); while the evaluation set consists of both \emph{known} and \emph{unknown} (S6-S10) attacks. 
More details and protocols about the ASVspoof database can be found in~\cite{wu15asvspoof}.

This noisy version aims to quantify the effects of current spoofing detection algorithms in noise conditions and to facilitate future assessment work in this task. In this section, we will briefly introduce the types of noise to be added, and the procedure of adding noise.

\subsection{Noise signals}

Five types of noise signals, representing the probable application scenarios, are used for the construction of the noisy ASVspoof database. A subset of three types of noise, white noise, speech babble and vehicle interior noise, are selected from NOISEX-92 database~\cite{Varga1992noisex}. Another two types mixed noise, street noise and cafe noise, are selected from QUT-NOISE database~\cite{dean2015QUT}. These are standard types of additive noise used for speech recognition~\cite{varga1993assessment, hirsch2000aurora, dean2015QUT}, speaker verification~\cite{drygajlo1998speaker, saeidi2010temporally} and speech enhancement~\cite{virag1999single}. We briefly describe these noises as follows:

\leftmargini=5mm
\begin{itemize}
	\item \textbf{White Noise}:  The random signal with a constant power spectral density.
	\item \textbf{Babble Noise}: Speech babble and the recording is made in a canteen with 100 people speaking. 
	\item \textbf{Volvo Noise}: Vehicle interior noise and the recording is made in Volvo 340 on an asphalt road, rainy conditions. 
	\item \textbf{Street Noise}: Mixed noise, which is made at the roadside near inner-city, mainly consisting of road traffic noise, pedestrian traffic noise and bird noise.
	\item \textbf{Cafe Noise}: Mixed noise, which is made in outdoor cafe environment, mainly consisting of speech babble and kitchen noise from the cafe environment. 	
\end{itemize}

\begin{figure}[!htb]
\centering
\includegraphics[width=8cm]{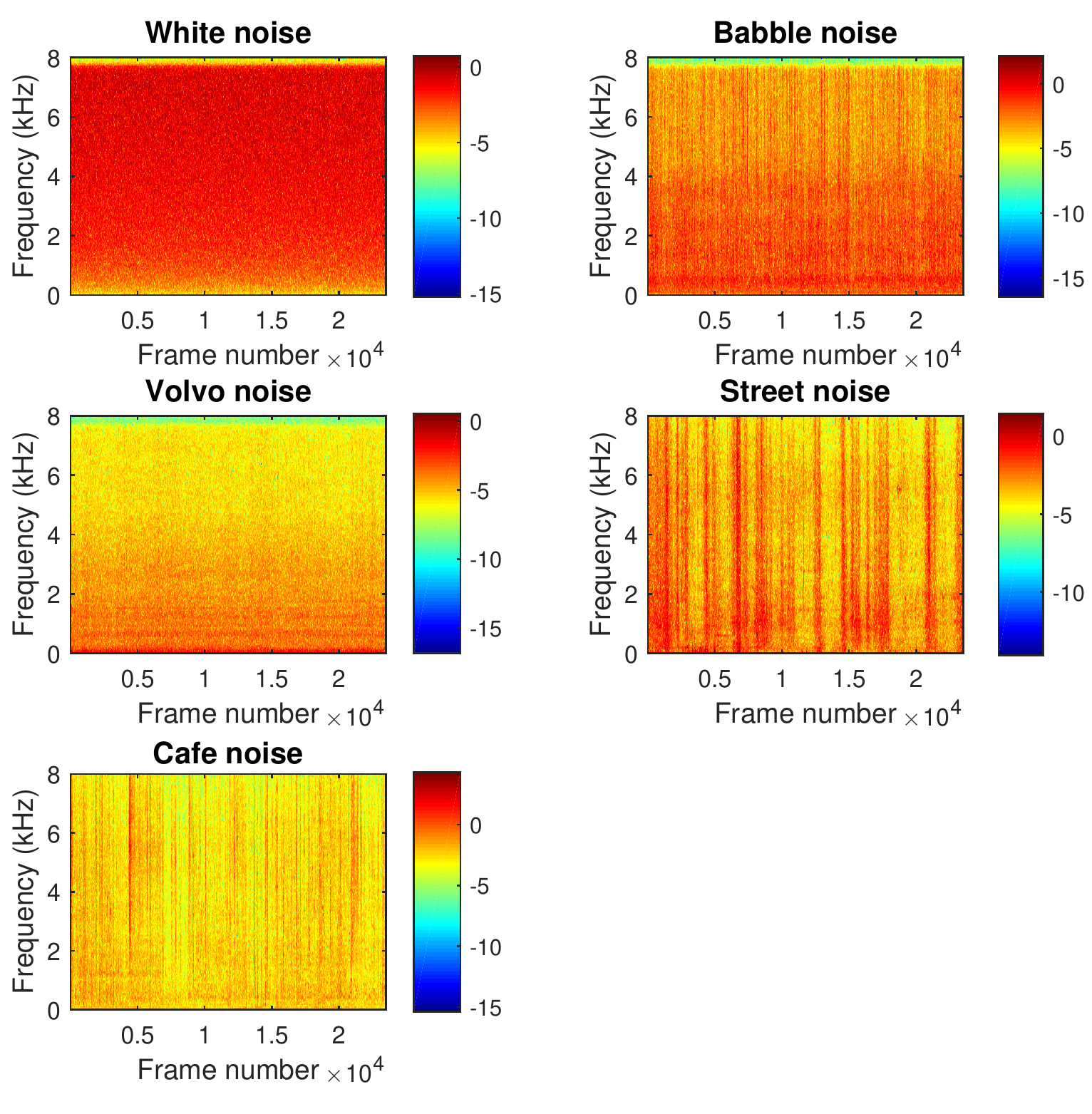}
\vspace{-2mm}
\caption{The spectrogram of different noise file.}
\label{fig:noise_sp}
\end{figure}

Figure~\ref{fig:noise_sp} shows the spectrogram of all the five noises. 
We can classify the noises into stationary noise and non-stationary noise. White noise and Volvo noise are stationary noise. While babble noise, street noise and cafe noise are recorded in non-stationary noise environment, whose the magnitude and phase spectrogram are varied over time. 

\subsection{Adding noise}
\label{sssec:add_noise}

The data from ASVspoof database are considered as clean data. Noise is artificially added to the clean data.
The Filtering and Noise Adding Tool (FaNT)\footnote{http://dnt.kr.hs-niederrhein.de/} is used for the adding noise process.
The noisy signals are generated by adding the clean speech and noise files together at various SNRs.
As the silence periods appear in many speech files of ASVspoof database, it is important to calculate the SNR only based on the sections of speech signal. 
The bandpass filter or frequency weighting are often used for SNR calculation, to ensure the SNR are appropriate and comparable.
In this work, to add the noise based on the human hearing perception, we define the SNR as the ratio of signal to noise energy after filtering both signals with the A-weighting filter. This filter emphasizes the frequencies around 3 kHz to 6 kHz where the human ear is most sensitive and give a lower response for the very high and very low frequencies to which the ear is less sensitive~\cite{fletcher1933loudness}. Hence, noises such as Volvo noise, whose energy distribution concentrates more on very low frequency (below 1 kHz), tend to have higher energy at the same SNR level.


For each clean signal in the development and evaluation sets of ASVspoof database, fifteen noisy versions of the signal are generated consisting of five types of noise in three SNR levels.
Given a clean signal, for each noise type, we take a segment of the noise signal with equal length as the clean signal but random starting point from the whole noise file. Then the noise segment is scaled and added to the clean signal in 20 dB, 10 dB and 0 dB SNR levels.

After the adding noise process, the clipping may occur, especially at low SNR levels due to high noise energy. 
In order to maintain a stable spectrogram representation of the signal, the signal is scaled to avoid the clipping.

All the five types of noise are added to the ASVspoof database at the SNRs of 20 dB, 10dB, and 0 dB. Hence, the ASVspoof noisy database is fifteen times of the clean database.

\section{Benchmarking system}

In order to demonstrate the utility of the ASVspoof noisy database, we conduct a series of experiments to examine the performance of spoofing speech detection system on a range of SNRs in all five noise scenarios.

\begin{figure}[!htb]
\centering
\includegraphics[width=7.5cm]{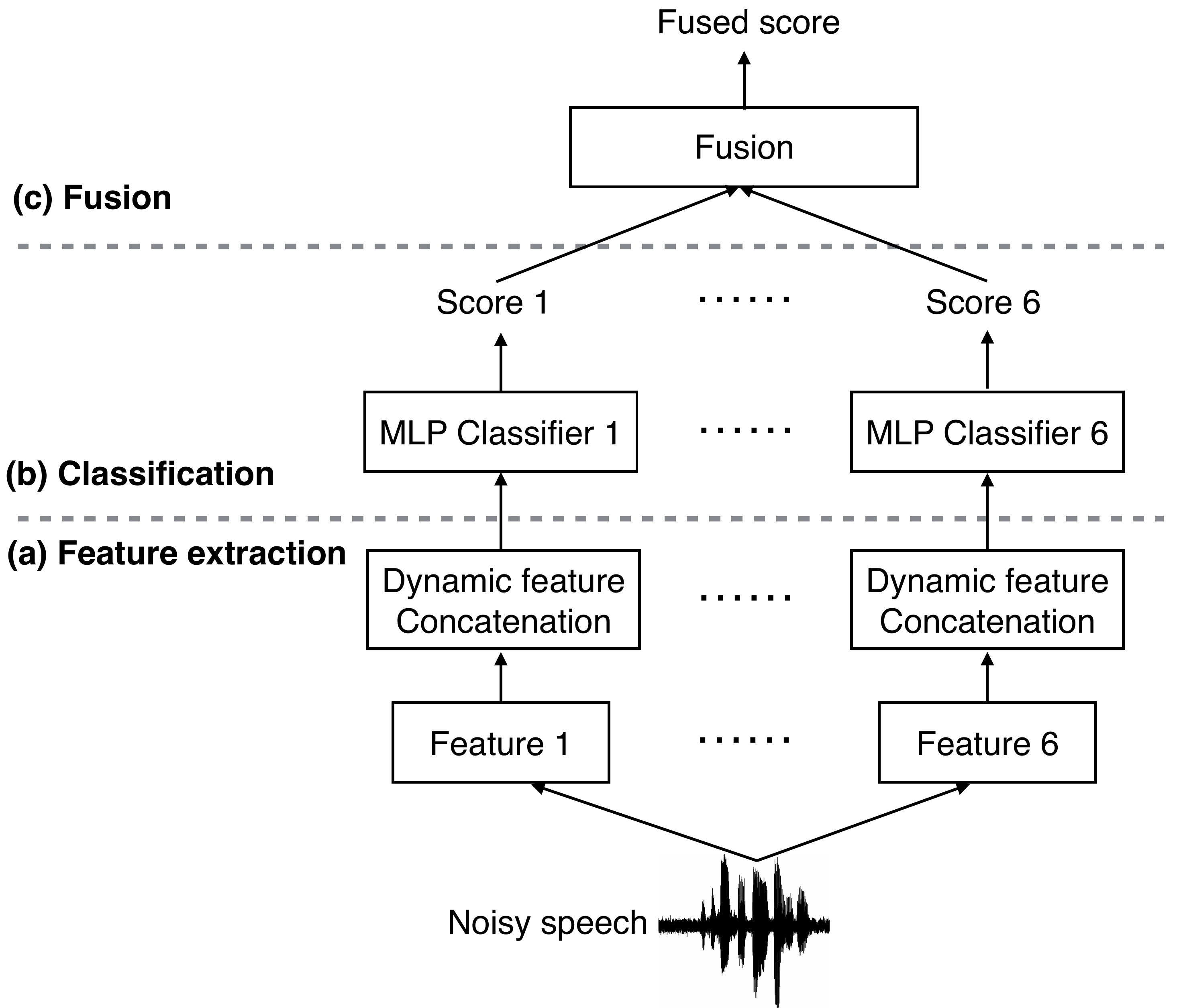}
\vspace{-2mm}
\caption{The architecture of our detection system, including (a) the feature extraction module, (b) the classification module and (c) the score fusion module.}
\label{fig:ASV_sys}
\end{figure}

The detection system, as shown in Figure~\ref{fig:ASV_sys}, consists of (a) the feature extraction module to extract six types of feature used for classification; (b) the classification module to calculate the score for each feature; (c) the score fusion module to fused the scores obtained from six classifiers. The details of these three modules are introduced as follows.

\subsection{Feature extraction}
\label{ssec:feature_extraction}

Similar to our previous system described in~\cite{xiao2015spoofing, tian2016spoof}, six types of feature are extracted. 
As shown in Figure~\ref{fig:ASV_sys} (a), given a noisy waveform, the Hamming window and direct current (DC) offset removal are applied on each analysis frame. Then short-time Fourier transform (STFT) is applied on the speech signal using analysis window of 25ms with 15ms overlap. The FFT length is chosen to be 512 and the dimension of all the original features are 256. 
For the $n$-th frame, the magnitude and phase spectrum, $|\mathrm{X}(n, \omega)|$ and $\theta(n, \omega)$, are obtained by 
\begin{equation}
\mathrm{X}(n, \omega) = |\mathrm{X}(n, \omega)|e^{j\theta(n, \omega)},
\label{(logMagEq)}
\end{equation}

After that, two magnitude-based features, namely log magnitude spectrum (LMS) and residual log magnitude spectrum (RLMS) are derived from magnitude spectrum. Four phase-based features, namely instantaneous frequency derivative (IF), baseband phase difference (BPD), group delay (GD) and modified group delay (MGD), are derived from phase spectrum.

\begin{figure*}[!htb]
\centering
\includegraphics[width=15cm]{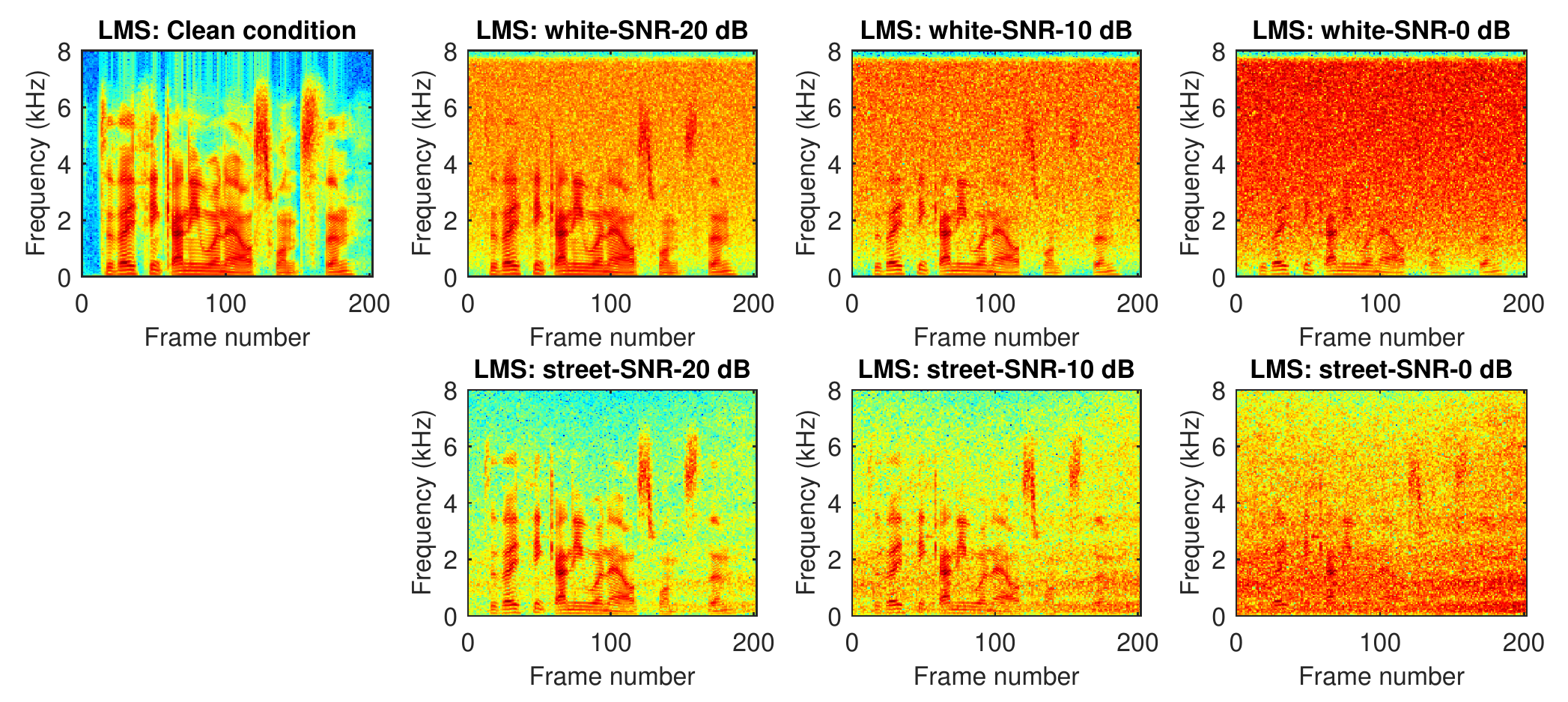}
\vspace{-2mm}
\caption{Demonstration of the LMS feature for utterance D14\_1000302, in both clean and noise scenarios.}
\label{fig:noisy_features_sp}
\end{figure*}

\begin{figure*}[!htb]
\centering
\includegraphics[width=15cm]{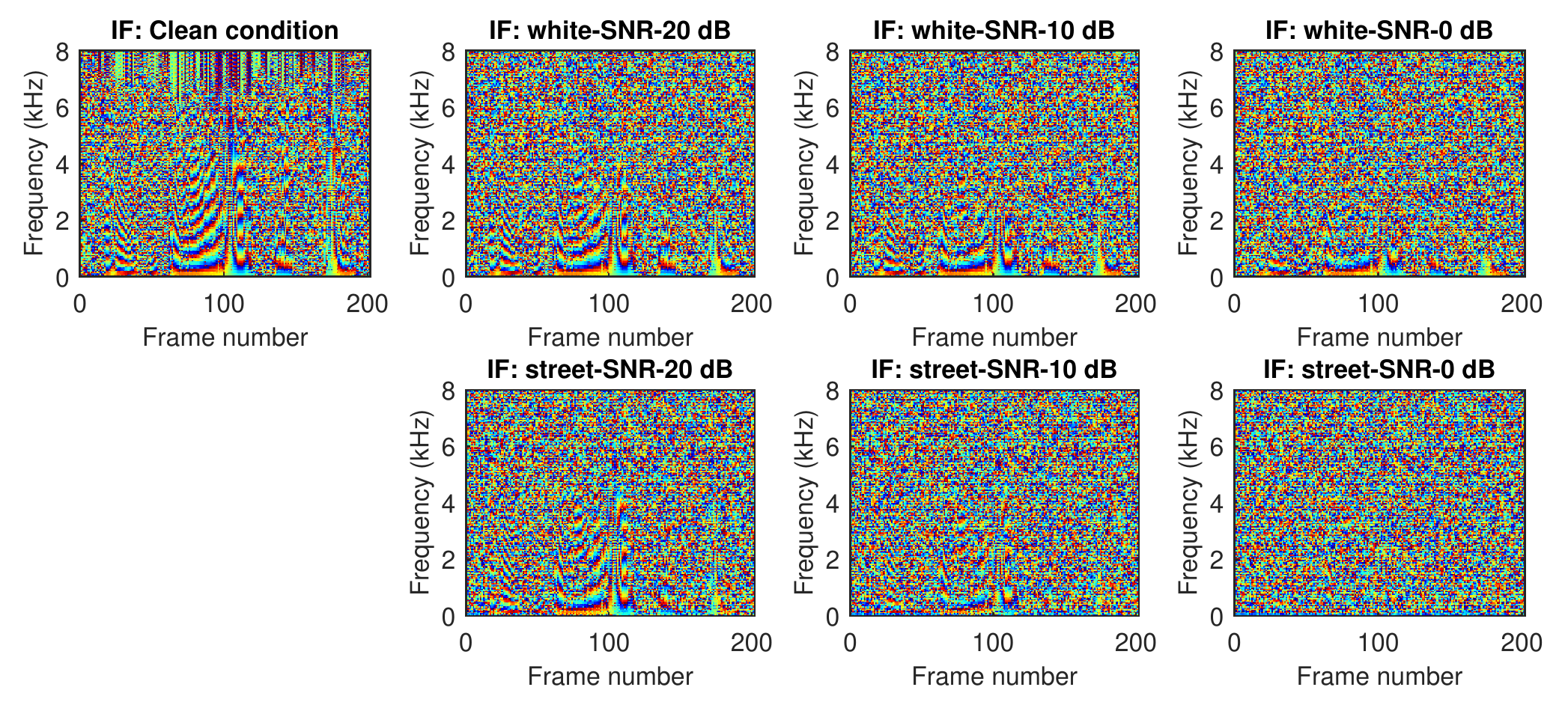}
\vspace{-2mm}
\caption{Demonstration of the IF features for utterance D14\_1000302, in both clean and noise scenarios.}
\label{fig:noisy_features_IF}
\end{figure*}

The features are summarized as follows:

\leftmargini=5mm
\begin{itemize}
	\item \textbf{LMS}: The log magnitude spectrum feature can be expressed as $\mathrm{LMS(n, \omega)}=log(|\mathrm{X}(n, \omega)|)$. The LMS contains the formant information, harmonic structure and all the spectral detail of speech signal. 
	\item \textbf{RLMS}: The residual log magnitude spectrum feature is the LMS extracted from the linear predictive coding (LPC) residual signal. As the formant information is removed, this feature can better analyses the harmonic structure and spectral details. 
	\item \textbf{IF}: Instantaneous frequency~\cite{alsteris2007short} is the derivative of the phase along time axis and captures the temporal information of phase. It is defined as:
\begin{align}
\mathrm{IF}(n, \omega) = \mathrm{princ} (\theta(n, \omega)-\theta(n-1, \omega)).
\end{align}
where $\mathrm{princ}(\cdot)$ represents the principal value operator, mapping the input onto $[-\pi; \pi]$ interval by adding integer numbers of $2\pi$. 
	\item \textbf{BPD}: Baseband phase difference~\cite{krawczyk2014stft} is another phase feature derived from IF and baseband STFT. For the $n$-th frame, the BPD is calculated as: 
	\begin{equation}
	\mathrm{BPD}(n, \omega) = \mathrm{princ}(\mathrm{IF}(n, \omega) - \Omega_kl),
	\end{equation}
    where $\Omega_k = 2 \pi k/L $ is the normalized angular frequencies. 
	\item \textbf{GD}: Group delay~\cite{yegnanarayana1992significance} is a representation of filter phase response, which is defined as the negative derivative of the Fourier transform phase. It is a frame-based feature, used to capture the phase distortion along frequency axis.
\begin{equation}
	\mathrm{GD}(n, \omega) = \mathrm{princ} \{\theta(n, \omega) - \theta(n, \omega-1)\},
\end{equation}
	\item \textbf{MGD}: A variation of GD, which can obtain a more clear phase pattern than GD. The MGD~\cite{yegnanarayana1992significance} feature of frame $n$ is calculated as:
	\begin{equation}
	\tau(n, \omega) = \frac{X_R(n, \omega)Y_R(n, \omega) + X_I(n, \omega)Y_I(n, \omega)}{|S(n, \omega)|^{2\gamma}},
	\end{equation}
	\begin{equation}
	\mathrm{MGD}(n, \omega) = \frac{\tau(n, \omega)}{|\tau(n, \omega)|} |\tau(n, \omega)|^\alpha,
	\label{eq:mgd}
	\end{equation}
where $X_R(n, \omega)$ and $X_I(n, \omega)$ denote the real and image part of STFT for $x(l)$; while $Y_R(n, \omega)$ and $Y_I(n, \omega)$ denote the real and image part of STFT for $lx(l)$, respectively. 
$S(n, \omega)$ is the smoothed spectrum of $|X(n, \omega)|$. 
Based on experimental results, the values of $\gamma$ and $\alpha$ are set to 0.7 and 0.2 respectively. 
\end{itemize}

In Fig~\ref{fig:noisy_features_sp} and Fig~\ref{fig:noisy_features_IF}, we present the LMS and IF features extracted from clean and the noisy signals. The noisy signals include the white and street noise scenarios at the SNR of 20 dB, 10 dB and 0 dB. We observe that both LMS and IF feature are distorted by additive noise significantly. The patterns become more blurred for lower SNRs.

\subsection{Classifier}
\label{ssec:classifier}
Figure~\ref{fig:ASV_sys} (b) shows the classification part of the detection system. Our previous multilayer perceptron (MLP) based spoofing speech detection system~\cite{tian2016spoof} is used in this work.
Each of the features mentioned above with its delta and acceleration coefficients is used as the input vector to train its own classifier. 
The MLP, which contains one hidden layer with 2,048 sigmoid nodes, is used to predict the posterior probability of the input vector being synthetic speech. 
The score is calculated by averaging the posterior probabilities of all the frames over the utterance.
Noted that, in this work, all the MLP classifier are trained from clean data.

\subsection{Evaluation metrics and fusion}
\label{ssec:fusion}

The equal error rate (EER), where false acceptance rate and miss rejection rate become equal, is used to evaluate the system performance.

As described in Section~\ref{ssec:feature_extraction}, different features are designed to detect different types of artifacts. In order to benefit the advantages of each feature and improve the system stability, a score level fusion is applied. Figure~\ref{fig:ASV_sys} (c) shows score fusion of the detection system. 

The fusion is applied on the feature-based results in different noise scenarios. 
Based on our preliminary experiments, the weighted summation fusion, tuned on the development data set, exhibits the over-fitting effects at street and cafe noise scenarios.
To avoid this problem, the scores of all systems are simply averaged to produce the final score.

In this work, the Bosaris toolkit\footnote{https://sites.google.com/site/bosaristoolkit/} is used to compute the EERs of each feature and the fused system.

\section{Experiments}

\subsection{Experimental setups}
\label{ssec:database}

The database used in the experiments consist of three subsets, including training set, development set and evaluation set. 
The training set is clean speech data taken from ASVspoof database. 
As the training set consists of clean data only, it models the speech without noise distortion and represents all the speech information. 
The best performance of the clean classifier is obtained in the case of testing on clean data, which can be found in our previous work~\cite{tian2016spoof}. 

The development and evaluation sets are chosen from the noisy ASVspoof database, including five noise scenarios at three different SNRs as described in Section~\ref{sec:noise_database}.
Because the classifier used in these experiments is the same as that of our previous work~\cite{tian2016spoof}, these results are comparable with the results in clean condition.

\subsection{Evaluation results}
\label{ssec:evaluation_results}

\begin{table*}[!t,h]
\caption{\it Average EERs (\%) of different features on the evaluation set. Clean indicates the results of our previous work~\cite{tian2016spoof}.}
\label{table:eval_results} \centering\footnotesize
  \includegraphics[width=17cm]{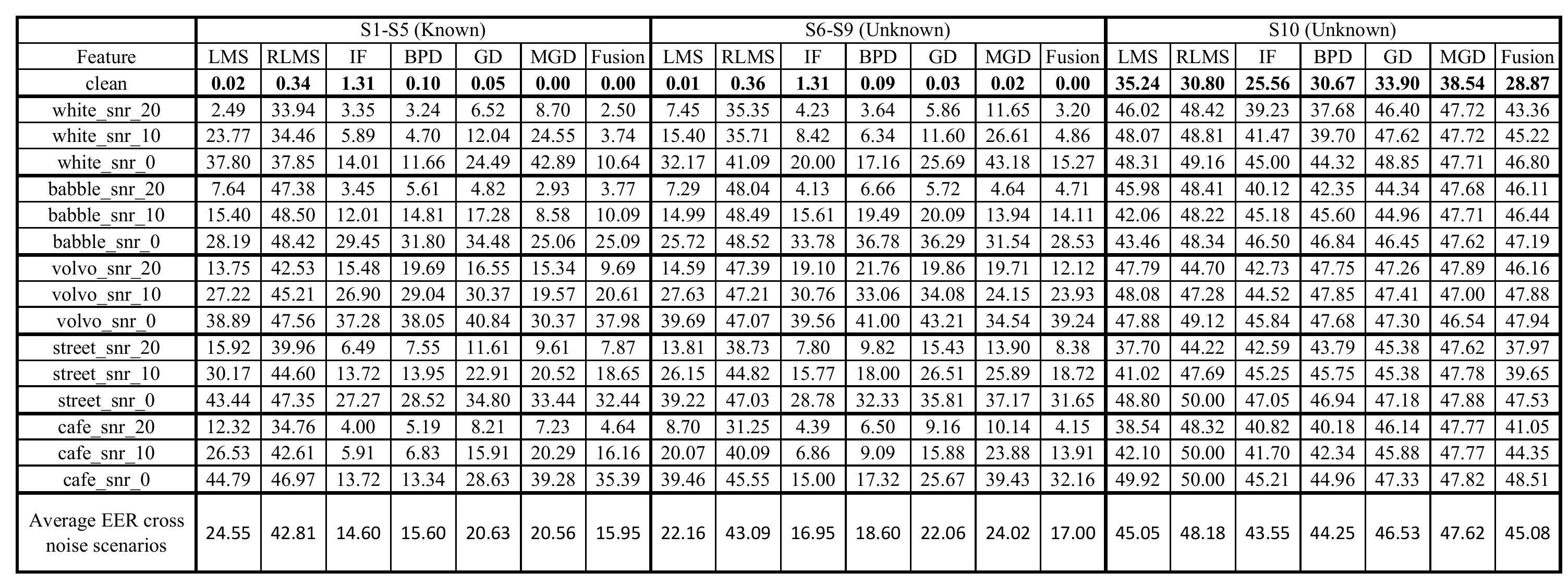}
\end{table*}

As the results on the development set are similar to that of the S1 to S5 on the evaluation set, only the feature-based results of the evaluation set are reported. The results are shown independently as the known attacks (S1-S5), the unknown attacks (S6-S9) and the unknown attacks generated by waveform concatenation (S10). The EERs are listed in Table~\ref{table:eval_results} for both noisy dataset and clean dataset using the clean classifier.

We first analyse the effect of noisy data for the detection system using different features.
In general, across all the five noise scenarios, the systems perform worse than that of clean condition. As expected, in most spoof attacks, the detection performance deteriorates as SNR decreases.
We notice that, in most noisy scenarios, the magnitude-based features, LMS and RLMS, perform worse than the phase-based features, IF, BPD, GD and MGD. In particular, in all the spoofing attacks, the RLMS obtains much higher EERs than other features. This may be due to that the LPC filter is not robust in noisy environments~\cite{kay1979effects}, which affects the quality of RLMS.
Among the phase-based features, IF and BPD outperform other features in terms of the average EERs over all the noise scenarios. 
We also found that, some features are effective for particular noise scenario. For example, in the babble noise scenario, the MGD is capable to obtain low error rates. While in white, street, and cafe noises, IF and BPD perform much better than other features.

Next, we compare the performance across different types of attack in noise conditions. 
Because S1-S5 attacks are available for training, even in noisy condition, the lower error rates are obtained in these attacks than the rest types of attack.
Although the error rates of S6-S9 is higher than that of S1-S5, the results still comparable. This is consistent with the results in clean condition~\cite{tian2016spoof}.
For S10, even apply the score fusion, the error rates of all the features are significantly higher than that of S1-S9. 
Hence, we conclude that in both clean and noisy conditions, the detection of S10 is still the most challenge task among the spoofing attacks. 

\subsection{Fusion results}
\label{ssec:fusion_results}

\begin{table*}[!t,h]
\caption{\it EERs (\%) of fused system on both development and evaluation sets. Clean indicates the results of our previous work~\cite{tian2016spoof}.}
\label{table:fusion_results} \centering\footnotesize
  \includegraphics[width=16cm]{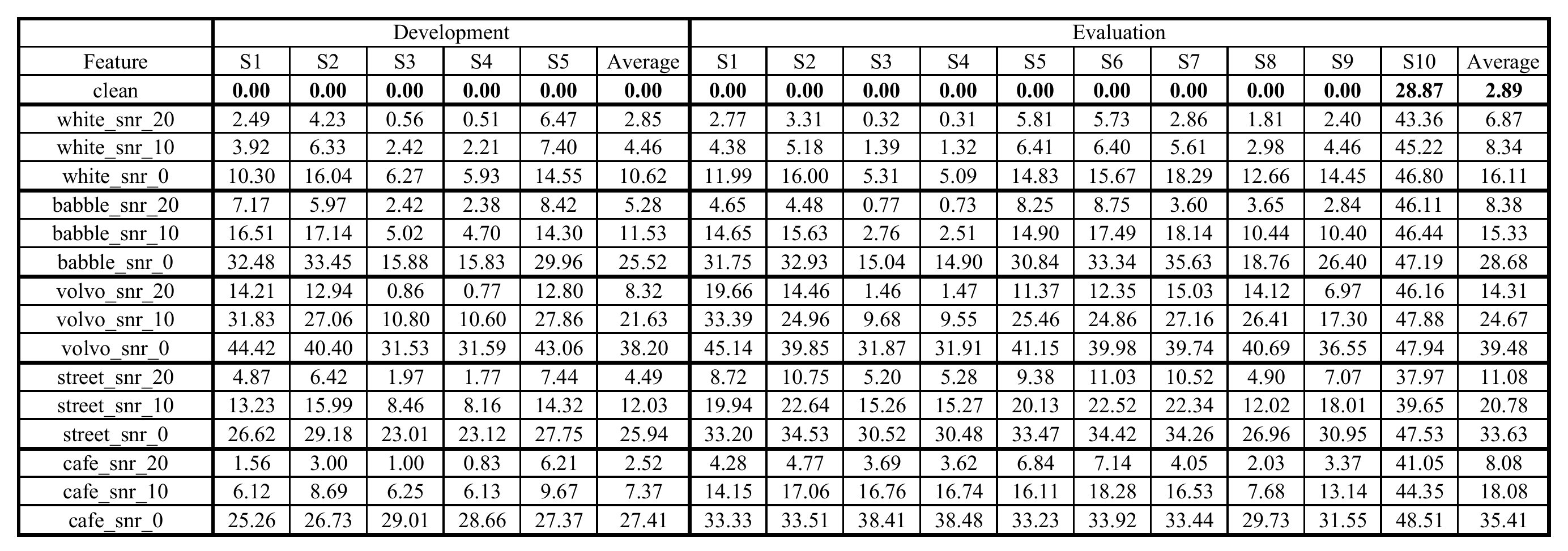}
\end{table*}

Table~\ref{table:fusion_results} presents the results of fused systems on both development and evaluation sets.
We first examine the results at different SNRs.
In all the noise scenarios at the SNR of 20 dB, the system performance degradation significant differ for different noise scenarios. On the development set, the EERs of different fuse systems are between 2.52\% (cafe noise) to 8.32\% (Volvo noise); while on the evaluation set, the EERs are varies between 5.25\% (white noise) to 14.31\% (Volvo noise).
However, at the SNR of 0 dB, all the systems performance degrade significantly on both development and evaluation set. Fig~\ref{fig:noisy_features_sp} and Fig~\ref{fig:noisy_features_IF} show that most of the feature patterns are lost in such low SNR.

Then, we analyse the fused results in different noise scenarios.
Among all the noise scenarios, the system under white noise scenario performs best, which constantly achieves lowest error rates.
Especially, at low SNRs, 10 dB and 0 dB, the system under white noise outperform that of other noise scenarios significantly.
As discussed in Section~\ref{sssec:add_noise}, compare to other types of noise, the Volvo noise tends to give higher energy at the same SNR level, resulting in more distortions in the noisy signals and higher EERs.

For the non-stationary noisy conditions, the features distorted by such noise are time-varying. Consequently, in these noise conditions, the system performance degrades more than white noise scenarios. 
This provides more challenge to the system to detect the spoofed attacks in such noisy conditions.

\section{Conclusions}

In this paper, we constructed a noisy database for spoofing and anti-spoofing research. This database is generated from the ASVspoof database, adding five types of additive noise at three SNR levels.
To provide the benchmark results, we use the state-of-the-art spoofing detection system to detect the spoofing attacks in noisy conditions.
The preliminary results using the classifier trained from clean data shown that,
\leftmargini=5mm
\begin{itemize}
\item the performance of detection systems degrade in all the noise scenarios. The system performance deteriorates as SNR decreases;
\item in noisy environments, even the best EER is about 4\% lower (white noise at 20 dB) than that of clean condition;
\item in general, the non-stationary noises affect the detection system more seriously;
\item the system performance varies significantly under different noise scenarios and the phase-based features are noise robust than magnitude-based features. 
\end{itemize}

In this paper, we only presented benchmark results to demonstrate the vulnerability of current spoofing detection systems under additive noise conditions.
In future work, we will use convolutional noise, such as channel noise and reverberate noise, to simulate the spoofing attacks in more complex noise scenarios.
Moreover, the classifier presented in this work was trained from the clean data.
There are also plans to exam the effectiveness of multi-condition training for spoofing detection under noisy conditions. 
Finally, a proper score fusion method will be investigated to both avoid the over-fitting effects and improve the system performance.


\section{Acknowledgment}
This research is supported by the National Research Foundation Singapore under its Interactive Digital Media (IDM) Strategic Research Programme.
This work is also supported by the DSO funded project MAISON DSOCL14045, Singapore.

\bibliographystyle{IEEEbib}
\bibliography{Odyssey2016_BibEntries}

\end{document}